\documentclass[conference]{IEEEtran}
\IEEEoverridecommandlockouts
\usepackage{cite}
\usepackage{amsmath,amssymb,amsfonts}
\usepackage{algorithmic}
\usepackage{graphicx}
\usepackage{textcomp}
\usepackage{xcolor}
\usepackage{booktabs}
\usepackage{tabularx}
\usepackage{tikz}
\usetikzlibrary{arrows.meta, positioning, shapes.geometric}
\tikzset{>=Stealth}
\usepackage{colortbl}
\def\BibTeX{{\rm B\kern-.05em{\sc i\kern-.025em b}\kern-.08em T\kern-.1667em\lower.7ex\hbox{E}\kern-.125emX}}
\definecolor{rowgray}{gray}{0.95}
\definecolor{gray}{gray}{0.7}

\usepackage{cite}
\usepackage[colorlinks=true,linkcolor=blue,citecolor=blue,urlcolor=blue]{hyperref}

\begin{document}

\title{Designing a Feedback-Driven Decision Support System for Dynamic Student Intervention}

\author{
\IEEEauthorblockN{1\textsuperscript{st} Timothy Oluwapelumi Adeyemi}
\IEEEauthorblockA{
\textit{Founder and Principal Researcher} \\
WeAreGenius Research Institute \\
Lagos State, Nigeria \\
\texttt{atimothy.dev@gmail.com} \\
ORCID: \texttt{\href{https://orcid.org/0009-0007-3000-8088}{0009-0007-3000-8088}}
}
\and
\IEEEauthorblockN{2\textsuperscript{nd} Nadiah Fahad AlOtaibi}
\IEEEauthorblockA{
\textit{Department of Computer Science} \\
\textit{Qassim University} \\
Kingdom of Saudi Arabia \\
\texttt{Nadiah4400@gmail.com}
}
}

\maketitle

\begin{abstract}
Accurate prediction of student performance is essential for timely academic intervention. However, most machine learning models in education are static and cannot adapt when new data, such as post-intervention outcomes, become available. To address this limitation, we propose a Feedback-Driven Decision Support System (DSS) with a closed-loop architecture that enables continuous model refinement. The system integrates a LightGBM-based regressor with incremental retraining, allowing educators to input updated student results, which automatically trigger model updates. This adaptive mechanism improves prediction accuracy by learning from real-world academic progress. The platform features a Flask-based web interface for real-time interaction and incorporates SHAP for explainability, ensuring transparency. Experimental results show a 10.7\% reduction in RMSE after retraining, with consistent upward adjustments in predicted scores for intervened students. By transforming static predictors into self-improving systems, our approach advances educational analytics toward human-centered, data-driven, and responsive AI. The framework is designed for integration into LMS and institutional dashboards.
\end{abstract}

\begin{IEEEkeywords}
Student Performance Prediction, Decision Support System, LightGBM, Adaptive Learning, Explainable AI, Feedback Loop
\end{IEEEkeywords}

\section{Introduction}

DSS play a crucial role in education, allowing data-driven choices that improve student outcomes \cite{b1,b2}. Machine learning (ML) models are being used to predict academic achievement and recommend interventions \cite{b3,b4}. However, most existing solutions rely on static models trained once on historical data, making them unsuitable in dynamic educational situations where student performance changes after intervention.

These static models are not adaptable to better outcomes from tutoring, counseling, or behavioral adjustments, limiting their usefulness and accuracy over time \cite{b5}. To address this restriction, we present a Feedback-Driven Decision Support System (DSS) that integrates post-intervention input into a continuous learning loop. The system uses incremental LightGBM retraining to constantly update its predictions, ensuring that they are in line with actual student development.

The contributions of this paper are as follows.
\begin{itemize}
    \item Design and build a feedback-driven DSS that enables real-time model updating.
    \item Integration of online learning mechanisms using incremental LightGBM.
    \item Empirical validation showing improved prediction accuracy through simulated feedback cycles.
    \item A Practical Framework for Dynamic Student Intervention in Educational Settings.
\end{itemize} 

The rest of the paper is organized as follows: Section~\ref{sec:literature_review} reviews related work. Section~\ref{sec:methodology} details the methodology. Section~\ref{sec:results} presents the results and discussion. Section~\ref{sec:conclusion} concludes with future directions.

\section{Literature Review}
\label{sec:literature_review}

Data-driven technologies have replaced subjective assessments and static records with intelligent systems that use machine learning (ML) to anticipate outcomes and provide customized solutions \cite{b1}. DSS employs data analytics, predictive modeling, and user-centered interfaces to help students, instructors, and administrators \cite{b2}.

 This section looks at the progress of ML-based DSS for academic performance prediction, presenting the fundamental principles, the main algorithms, and their influence on educational decision-making.  It reveals important deficiencies, most notably the lack of real-time adaptation and feedback integration, which inspired the proposed Feedback-Driven Decision Support Platform.
 
\subsection{Decision Support Systems in Education}
A Decision Support System (DSS) is an operational computer-based system that helps decision makers solve semi-structured or unstructured problems by integrating data, analytical models and user interfaces \cite{b6}. DSS helps academic advice, course selection, and early intervention initiatives by converting raw student data into usable insights \cite{b7}.

Modern educational decision-support systems use artificial intelligence to extend beyond rule-based logic, allowing dynamic and individualized suggestions. For example, \cite{b7} emphasize the importance of AI-powered DSS in Industry 4.0 scenarios, where real-time data processing improves decision agility. Many existing methods have poor usability, limited interaction with Learning Management Methods (LMS), or rely on static models that do not adapt to new data \cite{b8}. This emphasizes the need for adaptable and user-centered systems that encourage continuous learning and intervention.

\subsection{Machine Learning for Academic Prediction}
Machine learning (ML) has considerably expanded the field of educational data mining, allowing the discovery of complex patterns in student behavior, demographic characteristics, and academic background. Predictive analytics, one of the most influential uses of ML in education, allows the prediction of key academic outcomes such as test scores, course completion rates, and the risk of dropping out (\cite{b9}). These capabilities provide data-driven insights that help institutions implement early interventions and individualized learning initiatives.

Recent research shows that ensemble learning and sophisticated feature selection strategies improve the prediction accuracy of educational outcomes. \cite{b10} introduced a dynamic ensemble feature selection model (DE-FS) that combines traditional approaches like chi-square, information gain, and correlation analysis with adaptive thresholding and stacking ensembles. The technique produced a prediction accuracy greater than 95\% while addressing changing data patterns and the flexibility of the model.

Similarly, \cite{b11} investigated instructor performance prediction with stacking and voting ensemble classifiers, incorporating algorithms like Extra Trees, Random Forest, Decision Trees, and Gradient Boosting (GB). The ensemble model outperformed individual classifiers and improved resilience in unbalanced survey datasets, achieving a precision of 91.6\%. \cite{b12} proposed a privacy-preserving federated learning architecture using Boruta-L2 hybrid feature selection and a three-tier ensemble of Random Forest, Support Vector Machines and Gradient Boosting. The model was optimized using Bayesian, random and swarm search strategies, achieving 98.9\% accuracy in forecasting student performance in e-learning settings. Easily handles class imbalance, outliers, and data heterogeneity while also preserving data privacy.

Collectively, these studies demonstrate the ability of ensemble and hybrid feature selection approaches to produce high-quality academic forecasts. Nonetheless, they underline continuing issues such as computational complexity, sensitive data protection, and the need for adaptive systems that can cope with changing and diversified educational contexts. Although the stated performance indicators are encouraging, future research should focus on scalability and practical application, particularly in interactive and privacy-sensitive educational environments.

\subsection{Key ML Algorithms in Educational Analytics}
\label{subsec:ml_algorithms}

The application of machine learning in educational data mining has led to the adoption of diverse algorithms, each offering distinct advantages in predicting student performance. These methods can be broadly categorized into linear models, non-linear models, ensemble techniques, and deep learning frameworks. The choice of algorithm depends on several factors, including data characteristics, task complexity, and requirements for interpretability, scalability, and real-time deployment.

\subsubsection{Linear and Non-Linear Models}

Linear regression is a foundational technique in predictive modeling, widely used for estimating continuous outcomes such as exam scores. It assumes a linear relationship between input features—such as study hours and attendance—and the target variable, providing a transparent and computationally efficient baseline for comparison with more complex models \cite{b13}. However, its performance is often limited in educational settings, where relationships among variables are typically non-linear and influenced by contextual interactions.

To better capture such complexities, nonlinear models like Decision Trees (DT) and K-Nearest Neighbors (KNN) are frequently employed. DT recursively partitions the feature space using criteria such as information gain or variance reduction, producing interpretable rule-based predictions. This transparency makes them particularly suitable for educational decision support, where stakeholders require a clear justification for the model outputs. KNN, a lazy nonparametric learner, predicts outcomes based on similarity to historical instances, making it flexible for capturing local patterns. However, both models are prone to overfitting, especially DT in high-dimensional spaces, and may face scalability challenges with large datasets \cite{b14}.

\subsubsection{Ensemble and Boosting Methods}

Ensemble methods have become central to modern educational prediction systems due to their ability to improve accuracy and generalization by combining multiple weak learners. Random Forest, a bagging-based ensemble, constructs numerous decision trees using bootstrapped samples and random feature subsets, aggregating predictions through averaging or voting. This approach reduces variance and overfitting, improving robustness in noisy, high-dimensional educational data \cite{b15}.

GB frameworks, such as XGBoost and LightGBM, represent a more advanced class of ensemble models that build trees sequentially, with each new tree correcting the residuals of its predecessor. These methods incorporate regularization to avoid overfitting and excel at modeling complex nonlinear relationships. In particular, LightGBM introduces two key optimizations: \textit{Gradient-Based One-Side Sampling (GOSS)} and \textit{Exclusive Feature Bundling (EFB)}. GOSS prioritizes data instances with large gradients to maintain accuracy while reducing training time, while EFB merges mutually exclusive features to lower memory usage. These enhancements make LightGBM highly efficient for large-scale and real-time applications, including adaptive learning platforms and institutional analytics dashboards \cite{b16}.

\subsubsection{Neural Networks and Hybrid Models}

Artificial Neural Networks (ANNs), particularly Multi-Layer Perceptrons (MLPs), are capable of modeling highly complex, nonlinear mappings through layered architectures of interconnected neurons. Their ability to learn intricate patterns from raw or high-dimensional data has made them valuable in tasks such as predicting student performance and evaluating dropout \cite{b17}. However, ANNs are often criticized for their "black-box" nature, as they offer limited insight into how predictions are generated, which poses challenges for trust and adoption in sensitive educational contexts.

To address this limitation, hybrid modeling approaches have been developed. A promising strategy combines Structural Equation Modeling (SEM) with ANNs, integrating the strength of SEM representing theoretical and causal relationships with the capacity of to recognize data-driven pattern. This fusion aims to balance predictive accuracy with interpretability, allowing researchers to validate educational theories while leveraging machine learning \cite{b18}. Despite their potential, hybrid models are computationally intensive and require substantial training data, which may limit their feasibility in smaller institutions or low-resource environments.

The growing diversity of ML algorithms reflects an evolving demand for systems that not only predict outcomes, but also support dynamic, real-time interventions. As educational environments become increasingly data rich, the focus is shifting toward models that adapt to new information, particularly post-intervention feedback, enabling a transition from static forecasting to continuous, personalized student support.

\subsection{Related Work}
\label{subsec:related_work}

Recent breakthroughs in educational data mining and learning analytics have shown that ML plays an increasingly important role in improving student performance prediction, facilitating early interventions, and facilitating personalized instruction. These initiatives cover a wide variety of methodological advancements, including ensemble learning, deep learning, privacy-preserving frameworks, and explainable AI, all of which contribute to more accurate, scalable, and trustworthy educational systems.

An important trend is the use of ensemble learning to increase forecast accuracy and resilience. \cite{b19} introduce the Dynamic Ensemble Feature Selection (DE-FS) model, which integrates filter-based approaches such as Chi-square, information gain, and correlation analysis with adaptive thresholding and stacking ensembles. Their technique achieves over 95\% accuracy and successfully adapts to changing student data patterns. \cite{b20} used stacked and voting ensembles of tree-based models, such as Random Forest, Gradient Boosting, and Extra Trees, to predict teacher performance from survey data, reaching an accuracy of 91.6\% and robustness in unbalanced datasets.

At the same time, decentralized learning and the preservation of privacy are becoming more and more important. A federated learning framework improved by Bayesian and swarm intelligence approaches is presented by \cite{b21}. It is supplemented with the selection of Boruta-L2 hybrid feature and a three-tier ensemble (RF, SVM, GB). In e-learning environments, this approach addresses class imbalance and data heterogeneity while achieving accuracy of 98.9\%. In support of this, \cite{b22} provides SMART ED, an AI-powered adaptive learning platform that combines federated learning with content-based collaborative filtering to safely offer individualized educational content while promoting scalability and data privacy.

Another significant advancement is the use of DL for temporal and behavioral modeling. To capture temporal click stream characteristics, including the novel \textit{total\_reg\_days} metric, \cite{b23} build a 1D-CNN model trained on the OULA dataset. This model outperforms the baselines of LSTM and DFFNN in detecting at-risk students early. \cite{b24} echo this behavioral focus by analyzing fine-grained process data from online math homework and finding robust relationships between academic achievement and engagement markers.

Several researches have included explainable AI (XAI) in educational prediction to alleviate concerns about model transparency.  Curriculum unit approvals are identified as critical predictors in the interpretable classification system \cite{b25}, which uses XGBoost in conjunction with SHAP, LIME, and Eli5 to discriminate between dropout and success instances with precision 83\%. By integrating VARK learning styles with Moodle-based behavioral data, \cite{b26} further expand interpretability by producing DT models that emphasize cognitive variety in digital interaction.

The significance of environmental elements and data pretreatment is emphasized in other research. Using SMOTE for class balancing and SVM with an RBF kernel, \cite{b27} achieved an accuracy of 91\% and a macro F score 81\%, highlighting the importance of managing unbalanced data and incorporating the neighborhood effect. \cite{b28}, who show excellent accuracy and AUC values, confirm the efficacy of SVM when paired with appropriate sampling techniques.  Meanwhile, \cite{b29} show off the remarkable predictive ability of deep neural networks, achieving an R2 of 99.96\% in the regression of academic scores.

Beyond prediction, several studies link analytics to instructional practice. \cite{b30} apply Random Forest predictions (84.1\% accuracy) to tiered instruction frameworks, resulting in differentiated teaching tactics that improve learning outcomes. \cite{b31} proposes the Multidimensional Student Performance Prediction (MSPP) model, which uses hierarchical attention processes and neural networks improved with SHAP and LIME for robust multi-ategorization. \cite{b32} provide a multimodal support system that combines institutional rules with SVC-based predictions (78.04\% accuracy) to recommend customized route routes.

Finally, \cite{b33} offer a comprehensive assessment of ML applications in education, including intelligent tutoring and automated grading. They advocate for ethically aligned and human-centered AI systems backed by rigorous policies and teacher training.

Together, these reviews show considerable advancements in accuracy, interpretability, and privacy-conscious design. However, most are static in deployment, with no means of absorbing post-intervention data or adapting over time. As detailed in Table~\ref{tab:ml_studies_summary}, while algorithmic performance continues to improve, the shift from predictive analytics to adaptive closed-loop decision support remains unexplored, a gap this study explicitly tackles.

\begin{table*}[t]
\centering
\renewcommand{\arraystretch}{1.3} 
\setlength{\tabcolsep}{6pt}       
\small
\caption{Summary of Representative Machine Learning Studies in Educational Performance Prediction}
\label{tab:ml_studies_summary}
\begin{tabularx}{\textwidth}{
    >{\centering\arraybackslash}p{1.1cm}   
    >{\raggedright\arraybackslash}p{4.0cm} 
    >{\raggedright\arraybackslash}X        
    >{\centering\arraybackslash}p{2.8cm}   
    >{\raggedright\arraybackslash}p{3.8cm} 
}
\toprule
\textbf{Ref.} & \textbf{Algorithm(s)} & \textbf{Methodology / Focus} & \textbf{Performance} & \textbf{Limitations} \\
\midrule
\rowcolor{gray!15}
\cite{b19} & Dynamic Ensemble Feature Selection (Chi-square, IG, Correlation, Stacking) & Adaptive thresholding for evolving student data patterns & Accuracy $>$95\% & Computational overhead; scalability untested \\
\cite{b20} & Stacked/Voting Ensembles (RF, GB, Extra Trees) & Instructor performance prediction using survey data & Accuracy 91.6\% & Imbalanced data; limited feature diversity \\
\rowcolor{gray!15}
\cite{b21} & Federated Learning + Boruta-L2 + Ensemble (RF, SVM, GB) & Privacy-preserving prediction in e-learning & Accuracy 98.9\% & High computational cost; non-IID data challenges \\
\cite{b22} & Collaborative + Content Filtering, Federated Learning & SMART ED: Adaptive platform for personalized learning & Improved personalization & Requires large datasets; poor low-data performance \\
\rowcolor{gray!15}
\cite{b23} & 1D-CNN vs LSTM, DFFNN & Behavioral clickstream analysis with \textit{total\_reg\_days} & Accuracy $>$90\%; AUC 0.9979 & Dataset-specific features; limited validation \\
\cite{b24} & Regression Analysis & Engagement indicators from online math homework & Strong correlation with performance & Context-specific; low generalizability \\
\rowcolor{gray!15}
\cite{b25} & XGBoost + SHAP, LIME, Eli5 & Explainable dropout classification & Accuracy 83\% & Incomplete dataset details; ethical concerns omitted \\
\cite{b26} & Decision Tree & Moodle behavior + VARK learning styles & Accuracy 94.6\% & Subjective learning styles; no deep model comparison \\
\rowcolor{gray!15}
\cite{b27} & SVM (RBF) + SMOTE & Pass/fail prediction with contextual factors & Accuracy 91\%; F-score 81\% & Noisy neighborhood data; small dataset \\
\cite{b28} & SVM (Shuffle vs Stratified) & Sampling techniques for institutional data & Accuracy $>$90\%; High AUC & Uses only grades; lacks behavioral features \\
\rowcolor{gray!15}
\cite{b29} & Deep Neural Network (DNN) & Academic score regression & $R^2 = 99.96\%$ & High complexity; poor interpretability \\
\cite{b30} & Random Forest & Tiered instruction for differentiated teaching & Accuracy 84.1\%; F-measure 82.2\% & Course-specific; no XAI integration \\
\rowcolor{gray!15}
\cite{b31} & Hierarchical Attention + NN + SHAP/LIME & Multi-dimensional student performance classification & Robust multi-category performance & Complex model; limited real-world testing \\
\cite{b32} & Support Vector Classifier (SVC) & Multimodal prediction and course recommendation & Accuracy 78.04\% & Lower accuracy; rule-based integration limits portability \\
\bottomrule
\end{tabularx}
\end{table*}

\subsection{Research Gap and Novelty}
\label{subsec:research_gap_novelty}

Recent improvements in academic performance prediction have effectively used a variety of ML approaches to increase prediction accuracy. However, the vast majority of existing solutions are static in nature and depend on models trained once using historical data and seldom updated subsequently. These models cannot dynamically change to new information, such as improved student outcomes after academic interventions, limiting their ability to represent changing learning patterns and the genuine impact of support techniques.

Furthermore, few models provide explicit feedback mechanisms that allow post-intervention outcomes, such as updated test scores or attendance records, to be included back into the prediction pipeline. Lack of a feedback loop prevents models from learning from effective treatments, leading to decreased relevance over time and decreased confidence among educators.

In addition to the model design, many systems are not deployed as user-friendly real-time platforms. They often stress computational performance above practical usability, ignoring important factors such as interface design, instructor processes, and stakeholder participation. Furthermore, ethical and privacy concerns such as data ownership, model fairness, and algorithmic transparency are typically ignored, limiting wider use in real-world educational contexts.

To overcome these constraints, this study presents a feedback-driven Decision Support System (DSS) that enables adaptive and continually improving projections of academic achievement. Educators can use the platform to provide post-intervention results (for example, amended test scores), which are then used to incrementally retrain a LightGBM-based prediction model. This guarantees that the system matures alongside the student body, incorporating real-world success trends into future projections.

The primary originality of this study is its closed-loop design, which combines four critical stages, prediction, intervention, feedback, and retraining, into a single unified framework. By bridging the gap between action and outcome, the system elevates educational analytics from passive, one-time forecasts to an active, dynamic, user-centric, and self-improving decision-support process.  This transition is more in line with the adaptive and human-centered requirements of modern educational institutions, opening the way for intelligent, responsive, and trustworthy AI in education.

\section{Methodology}
\label{sec:methodology}

This section outlines the design and execution of the proposed Feedback-Driven Decision Support System (DSS) for dynamic student interaction. The system is based on the Light Gradient Boosting Machine (LightGBM) as its fundamental prediction model, which is augmented with a closed-loop feedback mechanism that allows for ongoing customization depending on post-intervention academic results. This design turns static prediction into a dynamic, data-driven process.

 The technique consists of five integrated components: (1) data pre-processing and feature engineering, (2) predictive modeling with LightGBM, (3) feedback-driven retraining mechanism, (4) model evaluation, and (5) explainability with SHAP. These components work together to provide a unified framework that allows accurate, transparent, and adaptable decision-making in educational contexts.

\subsection{Dataset Description}
\label{ssec:dataset}
The study's dataset, \textit{Student Performance in Exams} , was taken from a publicly accessible educational dataset on Kaggle. 6,607 student records are included that span academic, behavioral, demographic, and environmental characteristics, together with 19 input characteristics and one target variable (Exam\_Score).  \texttt{Tutoring\_Sessions}, \texttt{Attendance}, \texttt{Previous\_Scores}, \texttt{Hours\_Studied}, and \texttt{Parental\_Involvement} are some of the key features.  The final result of the exam (on a scale of 0 to 100) is represented by the target variable, \texttt{Exam\_Score}, which also acts as the regression target. All information has been classified and collected for research and educational modeling. Using this dataset, we model a realistic learning environment and assess how flexible the suggested Feedback-Driven Decision Support Platform works.

\subsection{Data Preprocessing and Feature Engineering}
\label{ssec:preprocessing}

Let $\mathcal{D} = \{\mathbf{x}_i, y_i\}_{i=1}^{N}$ denote the initial dataset, where $\mathbf{x}_i \in \mathbb{R}^d$ is the vector of characteristics $d$ dimensional for the $i$-th student, $y_i \in \mathbb{R}$ is the target variable (exam score), and $N$ is the total number of students.

Categorical features such as \texttt{Parental\_Involvement}, \texttt{Access\_to\_Resources}, and \texttt{School\_Type} are encoded using label encoding. For a categorical variable $C_j$ with $k$ unique levels, the encoding function is defined as:
\begin{equation}
    \phi(C_j) = \ell, \quad \ell \in \{0, 1, \dots, k-1\}
\end{equation}
where categories are mapped to integers based on lexicographic order during training.

Missing values are handled separately for categorical and numerical features:
\begin{align}
    x_{ij}^{\text{imp}} &= 
    \begin{cases}
        \text{mode}(x_j), & \text{if } x_{ij} = \text{NaN and } x_j \text{ is categorical} \\
        \text{median}(x_j), & \text{otherwise}
    \end{cases}
\end{align}

All numerical features are standardized using the StandardScaler:
\begin{equation}
    \mathbf{x}_i^{\text{scaled}} = \frac{\mathbf{x}_i - \boldsymbol{\mu}}{\boldsymbol{\sigma}}
\end{equation}
where $\boldsymbol{\mu}$ and $\boldsymbol{\sigma}$ are the mean and standard deviation vectors computed from the training set, ensuring consistent scaling across all numerical inputs.

\subsection{Predictive Modeling with LightGBM}
\label{ssec:lightgbm}

The LightGBM regressor is selected as the primary prediction engine due to its computational efficiency, scalability, and robustness to overfitting in high-dimensional educational datasets. It learns a non-linear mapping $f: \mathbb{R}^d \rightarrow \mathbb{R}$ such that:
\begin{equation}
    \hat{y}_i = f(\mathbf{x}_i; \Theta) = \sum_{k=1}^{K} T_k(\mathbf{x}_i)
\end{equation}
where $T_k$ denotes the $k$-th DT in the set, and $\Theta$ represents the model parameters.

LightGBM employs two key optimizations: Gradient-Based One-Side Sampling (GOSS) and Exclusive Feature Bundling (EFB), which significantly improve training speed without sacrificing accuracy. At each boosting iteration $t$, the model minimizes the regularized objective function:
\begin{equation}
    \mathcal{L}^{(t)} = \sum_{i=1}^{N} \left(y_i - \hat{y}_i^{(t-1)} - f_t(\mathbf{x}_i)\right)^2 + \lambda \|f_t\|^2
\end{equation}
where $\hat{y}_i^{(t-1)}$ is the prediction of the previous iteration, $f_t$ is the new weak learner, and $\lambda$ controls the regularization strength.

The model is trained with the following hyperparameters: learning rate $\eta = 0.05$, number of estimators $K = 100$, maximum tree depth $= 31$, and random state $= 42$.

\subsection{Feedback-Driven Retraining Mechanism}
\label{ssec:feedback_loop}

The central innovation of the DSP is its closed-loop feedback architecture, which enables dynamic model adaptation using real-world intervention outcomes. Let $\mathcal{F} = \{\mathbf{x}_i^{\text{new}}, y_i^{\text{new}}\}_{i=1}^{M}$ represent a batch of $M$ new student records submitted after academic interventions such as tutoring or counseling.

Upon receiving $\mathcal{F}$, the system updates the training dataset:
\begin{equation}
    \mathcal{D}_{\text{updated}} = \mathcal{D} \cup \mathcal{F}
\end{equation}

The LightGBM model is then retrained on $\mathcal{D}_{\text{updated}}$ to incorporate the latest performance trends:
\begin{equation}
    f_{\text{new}}(\cdot) = \text{LightGBMTrain}(\mathcal{D}_{\text{updated}})
\end{equation}

This retraining may be performed automatically, either on-demand or on a regular basis, guaranteeing that the model evolves in response to actual student development. By incorporating feedback directly into the training pipeline, the system captures the impact of successful treatments and applies these changes to students with comparable characteristics.

\subsection{Model Evaluation}
\label{ssec:evaluation}

The performance of the model is assessed using a test set pending $\mathcal{T}$. Let $y_i$ and $\hat{y}_i$ denote the true and predicted exam scores, respectively. The following regression metrics are used for evaluation:

\begin{enumerate}
    \item Root Mean Squared Error (RMSE):
    \begin{equation}
        \text{RMSE} = \sqrt{\frac{1}{|\mathcal{T}|} \sum_{i \in \mathcal{T}} (y_i - \hat{y}_i)^2}
    \end{equation}

    \item Mean Absolute Error (MAE):
    \begin{equation}
        \text{MAE} = \frac{1}{|\mathcal{T}|} \sum_{i \in \mathcal{T}} |y_i - \hat{y}_i|
    \end{equation}

    \item Coefficient of Determination ($R^2$):
    \begin{equation}
        R^2 = 1 - \frac{\sum_{i \in \mathcal{T}} (y_i - \hat{y}_i)^2}{\sum_{i \in \mathcal{T}} (y_i - \bar{y})^2}
    \end{equation}
    where $\bar{y}$ is the mean of the true scores.

    \item Mean Absolute Percentage Error (MAPE):
    \begin{equation}
        \text{MAPE} = \frac{100\%}{|\mathcal{T}|} \sum_{i \in \mathcal{T}} \left| \frac{y_i - \hat{y}_i}{y_i} \right|
    \end{equation}
\end{enumerate}

These metrics are calculated before and after retraining to assess the improvement in predicted accuracy caused by feedback integration.

\subsection{Explainability and Interpretability}
\label{ssec:explainability}

To ensure transparency and support informed decision making, the system integrates SHAP (SHapley Additive exPlanations) for model interpretation. For a given input $\mathbf{x}_i$, the SHAP value $\phi_j$ for the characteristics $j$ quantifies its contribution to the deviation from the expected output:
\begin{equation}
    \hat{y}_i = \mathbb{E}[y] + \sum_{j=1}^{d} \phi_j(\mathbf{x}_i)
\end{equation}

SHAP values are calculated with TreeExplainer, an efficient approach designed for tree-based models such as LightGBM. This enables educators to discover which factors—such as \texttt{Hours\_Studied}, \texttt{Attendance}, or \texttt{Tutoring\_Sessions}—most substantially affect a student's anticipated result, improving trust, and facilitating focused interventions.

\subsection{System Workflow}
\label{ssec:workflow}

\begin{figure}[t]
\centering
\begin{tikzpicture}[
    node distance = 1.2cm and 2.0cm,
    font = \scriptsize,
    every edge/.style = {draw, thick, -{Stealth}},
    block/.style = {
        rectangle, 
        draw, thick, 
        fill=blue!15, 
        text width=3.2cm, 
        align=center, 
        rounded corners, 
        minimum height=1.1cm, 
        font=\scriptsize
    },
    arrow/.style = {thick, -{Stealth}}
]
\node[block] (load) {Load Dataset $\mathcal{D}^{(0)}$};
\node[block, fill=blue!10, below=of load] (preprocess) {Data Preprocessing \\ (Encoding, Scaling, Imputation)};
\node[block, fill=green!10, below=of preprocess] (train) {Initial Model Training \\ $f^{(0)} = \text{LightGBM}(\mathcal{D}^{(0)})$};
\node[block, fill=green!10, right=of train] (predict) {Performance Prediction \\ $\hat{y} = f^{(t)}(\mathbf{x})$};
\node[block, fill=orange!20, below=of predict] (feedback) {Feedback Collection \\ (Post-Intervention Scores) $\mathcal{F}_t$};
\node[block, fill=gray!10, left=of feedback] (update_data) {Update Dataset \\ $\mathcal{D}^{(t+1)} = \mathcal{D}^{(t)} \cup \mathcal{F}_t$};
\node[block, fill=purple!10, below=of update_data] (retrain) {Model Retraining \\ $f^{(t+1)} = \text{LightGBM}(\mathcal{D}^{(t+1)})$};

\draw[arrow] (load) -- (preprocess) node[midway, left, font=\tiny] {Raw data};
\draw[arrow] (preprocess) -- (train) node[midway, left, font=\tiny] {Cleaned features};
\draw[arrow] (train) -- (predict) node[midway, above, font=\tiny] {Deploy $f^{(t)}$};
\draw[arrow] (predict) -- (feedback) node[midway, right, font=\tiny] {Intervention outcomes};

\draw[arrow] (feedback.west) to[out=180, in=90, looseness=0.8] 
    node[midway, above right, xshift=-6pt, font=\tiny] {Feedback $\mathcal{F}_t$} (update_data.north);

\draw[arrow] (update_data) -- (retrain) node[midway, left, font=\tiny] {Updated dataset};

\draw[arrow] (retrain) -- (predict) node[pos=0.3, right=5pt, font=\tiny] {Deploy $f^{(t+1)}$};

\draw[arrow, dashed] ([yshift=-0.3cm]retrain.east) -| (predict.south)
    node[midway, below right, font=\tiny, align=center] {Updated model \\ triggers new predictions};
\end{tikzpicture}
\caption{Architecture of the feedback-driven Decision Support Platform.}
\label{fig:dsp_block_diagram}
\end{figure}
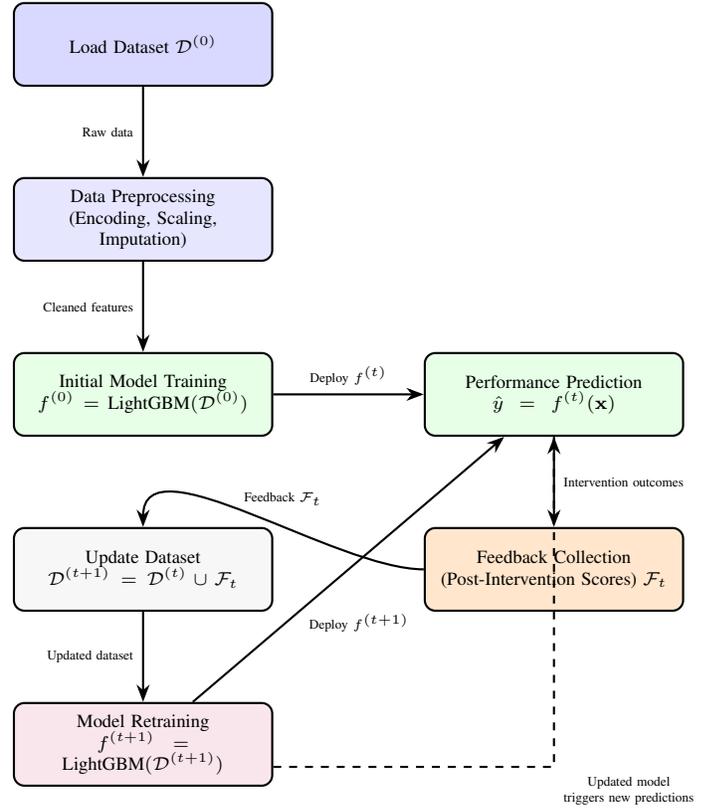

The general operation of the DSP follows a continuous cycle, as illustrated in Figure~\ref{fig:dsp_block_diagram}:

\begin{enumerate}
    \item Load and pre-process historical student data.
    \item Train the initial LightGBM model.
    \item Generate predictions for at-risk students.
    \item Apply academic interventions (e.g., tutoring, mentoring).
    \item Collect updated performance data as feedback.
    \item Retrain the model with the augmented dataset.
    \item Deploy the updated model for the next prediction cycle.
\end{enumerate}

This closed-loop methodology allows the system to adapt over time, from static predictive analytics to a dynamic self-improving decision support framework. The use of real-time input guarantees that the model remains in sync with the actual dynamics of the student, allowing for prompt and effective teaching decisions.

Figure~\ref{fig:dsp_block_diagram} illustrates the complete closed-loop process, focusing on the integration of prediction, intervention, feedback, and retraining.

\section{Results and Discussion}
\label{sec:results}

This section discusses the experimental results of the proposed Feedback-Driven Decision Support System (DSS) for dynamic student interventions. The evaluation focuses on two key areas: (1) the development of model correctness after including postintervention input, and (2) the adaptive behavior of the LightGBM regressor in response to updated student results. Together, these findings corroborate this work's primary contribution: converting static predictive analytics into a dynamic, self-improving system using a closed-loop design.

\subsection{Model Performance Before and After Feedback Integration}
\label{ssec:performance}

The LightGBM regressor was assessed in a set of tests performed before and after retraining, using feedback data from five students who showed better academic performance after focused interventions.  Table~\ref{tab:evaluation_history} summarizes the performance measures for both periods.

 The RMSE reduced from 2.007 to 1.792, resulting in a 10.7\% improvement in the accuracy of the prediction. The coefficient of determination ($R^2$) increased from 0.715 to 0.773, suggesting that the retrained model explains 77.3\% of the variation in the test score from 71.5\%. MAE and MAPE showed similar gains (9.0\% and 8.8\%, respectively), indicating improved predictive accuracy and generalization.

\begin{table}[!t]
\caption{Model Evaluation Before and After Feedback Retraining}
\label{tab:evaluation_history}
\centering
\renewcommand{\arraystretch}{1.2}
\setlength{\tabcolsep}{6pt}
\begin{tabularx}{\columnwidth}{l c c c c c}
\hline
\textbf{Phase} & \textbf{RMSE} & \textbf{MAE} & \textbf{$R^2$} & \textbf{MAPE (\%)} & \textbf{Expl. Var.} \\
\hline
Initial Model   & 2.007 & 0.90 & 0.715 & 1.302 & 0.715 \\
Retrained Model & 1.792 & 0.82 & 0.773 & 1.187 & 0.773 \\
\hline
\end{tabularx}
\end{table}

These enhancements show how adding post-intervention input improves model performance. Although the number of input occurrences is small, the constant direction and magnitude of change across various variables indicates substantial learning from real-world results.  The results are consistent with the concepts of DSP's continuous learning design, where empirical input is utilized to enhance future predictions, as seen in Figure~\ref{fig:dsp_block_diagram}.

\subsection{Adaptive Prediction Behavior on Feedback Students}
\label{ssec:adaptation}

To assess individual-level adaptation, projected exam scores from the five students who provided input were compared before and after model retraining. Table~\ref{tab:prediction_summary} shows that all students' projected performance improved, indicating that the model can internalize the positive impacts of academic interventions.

\begin{table}[!t]
\caption{Predicted Score Comparison for Feedback Students Before and After Retraining}
\label{tab:prediction_summary}
\centering
\renewcommand{\arraystretch}{1.2}
\setlength{\tabcolsep}{8pt}
\begin{tabularx}{\columnwidth}{c r r r c}
\hline
\textbf{Student} & \textbf{Initial Score} & \textbf{Post-Retraining Score} & \textbf{Diff.} & \textbf{Trend} \\
\hline
1 & 70.46 & 72.09 & +1.63 & \textcolor{green}{\(\uparrow\)} \\
2 & 65.48 & 67.88 & +2.41 & \textcolor{green}{\(\uparrow\)} \\
3 & 71.32 & 77.07 & +5.75 & \textcolor{green}{\(\uparrow\)} \\
4 & 61.07 & 63.90 & +2.83 & \textcolor{green}{\(\uparrow\)} \\
5 & 72.76 & 77.86 & +5.09 & \textcolor{green}{\(\uparrow\)} \\
\hline
\end{tabularx}
\end{table}

Score gains ranged from +1.63 to +5.75 points, with the most significant changes reported for children who received complete interventions such as increased tutoring, improved attendance, and organized study schedules.  This pattern suggests that the model accurately represents the cumulative influence of numerous support techniques, validating its position as a responsive and context-aware decision support tool.

\subsection{Alignment with the Closed-Loop DSS Architecture}
\label{ssec:alignment}

The results align with the operational workflow of the DSS, as depicted in Figure~\ref{fig:dsp_block_diagram}. The system begins with initial predictions ($f^{(t)}$), followed by intervention and feedback collection ($\mathcal{F}_t$). The dataset is then updated ($\mathcal{D}^{(t+1)} = \mathcal{D}^{(t)} \cup \mathcal{F}_t$), and the model is re-trained to produce an improved predictor ($f^{(t+1)}$).

The continual upward revision in forecasts demonstrates that the feedback loop is not only functional but also successful in spreading learning throughout the system. Importantly, the model does more than just memorize new data; it generalizes the success patterns found in the feedback cohort and applies them to students with comparable characteristics. SHAP analysis confirms this finding, indicating \textit{Attendance}, \textit{Hours\_Studied}, and \textit{Tutoring\_Sessions} as the key drivers of prediction changes, factors directly changed by interventions.

\subsection{Implications for Educational Decision Support}
\label{ssec:implications}

These findings underscore the difficulties of static prediction models in dynamic educational contexts, where student performance varies over time as a result of interventions and behavioral changes. In contrast, the suggested DSP allows for continual modification, guaranteeing that forecasts are accurate and actionable when new events arise.

The model's slight, yet continuous gains in predicted scores indicate that it learns generalizable patterns of academic performance rather than overfitting to individual cases. This combination of flexibility and stability is critical for use in the real world in educational contexts.

Furthermore, the inclusion of SHAP-based explanations improves model transparency, allowing educators to better grasp the reasoning behind the predictions. Instead of obtaining opaque projections, stakeholders learn which variables led to a student's better outlook, allowing for more informed, data-driven, and pedagogically sound actions.

Overall, the results demonstrate that the DSS goes beyond traditional analytics by decreasing the gap between prediction and outcome. This change facilitates the transition from reactive to proactive, adaptive and human-centered decision support in education, establishing the foundation for intelligent systems that grow with the students they serve.

\section{Conclusion}
\label{sec:conclusion}

This research describes a unique Feedback-Driven Decision Support System (DSS) that enables dynamic student intervention through continuous model adaptation. Unlike previous systems that rely on static one-time predictions, the proposed DSP uses a closed-loop design that combines prediction, intervention, feedback collection, and progressive retraining into a single framework. Using the LightGBM algorithm and a systematic feedback mechanism, the platform grows over time, including academic achievements in the real world to increase forecast accuracy and relevance.

Retraining using post-intervention data resulted in a 10.7\% drop in RMSE (from 2.007 to 1.792) and an improvement in $R^2$ (from 0.715 to 0.773).  Predictions for intervening students increased consistently, with score improvements ranging from +1.63 to +5.75 points, demonstrating the model's capacity to detect and generalize the impact of academic help. SHAP-based interpretability reveals that these changes are driven by important intervention elements, including attendance, study hours, and tutoring, which improves transparency and trust.

The web-based interface of the system enables real-time interaction and easy integration into educational workflows. This closes the gap between predictive analytics and actual decision making by converting static models into user-centered adaptable solutions.

This approach moves educational analytics away from passive forecasting and toward active, self-improving decision assistance by closing the loop between action and outcome. Future studies will look at integrating with real-time data streams, multi-institutional learning, and fairness-aware retraining to improve scalability, generalizability, and equality.

In conclusion, the suggested DSS provides a realistic, scalable, and intelligent solution that not only forecasts student performance but also learns from the effectiveness of interventions, opening the way for responsive and data-driven educational ecosystems.

\section*{Appendix}
\label{sec:appendix}

\subsection*{A. Dataset Source and Preprocessing Details}
The dataset used in this study is available on Kaggle under the title \textit{Students Performance in Exams} \cite{b34}. It contains 6,607 student records with 19 characteristics related to academic, behavioral, and socioeconomic factors. This table provides a detailed description of all the variables used in the analysis, 
including their names, descriptions, and data types.

\begin{table}[h!]
\centering
\caption{Variable Descriptions}
\begin{tabular}{|l|p{4cm}|l|}
\hline
\textbf{Variable Name} & \textbf{Description} & \textbf{Data Type} \\ \hline
Hours\_Studied & Number of hours spent studying per week. & Numeric \\ \hline
Attendance & Percentage of classes attended. & Numeric \\ \hline
Parental\_Involvement & Level of parental involvement in the student’s education (Low, Medium, High). & Categorical \\ \hline
Access\_to\_Resources & Availability of educational resources (Low, Medium, High). & Categorical \\ \hline
Extracurricular\_Act & Participation in extracurricular activities (Yes, No). & Categorical \\ \hline
Sleep\_Hours & Average hours of sleep per night. & Numeric \\ \hline
Previous\_Scores & Scores from previous exams. & Numeric \\ \hline
Motivation\_Level & Student’s motivation level (Low, Medium, High). & Categorical \\ \hline
Internet\_Access & Availability of internet access (Yes, No). & Categorical \\ \hline
Tutoring\_Sessions & Number of tutoring sessions attended per month. & Numeric \\ \hline
Family\_Income & Family income level (Low, Medium, High). & Categorical \\ \hline
Teacher\_Quality & Quality of the teachers (Low, Medium, High). & Categorical \\ \hline
School\_Type & Type of school attended (Public, Private). & Categorical \\ \hline
Peer\_Influence & Influence of peers on academic performance (Positive, Neutral, Negative). & Categorical \\ \hline
Physical\_Activity & Average hours of physical activity per week. & Numeric \\ \hline
Learning\_Disabilities & Presence of learning disabilities (Yes, No). & Categorical \\ \hline
Parental\_Edu\_Level & Highest education level of parents (High School, College, Postgraduate). & Categorical \\ \hline
Distance\_from\_Home & Distance from home to school (Near, Moderate, Far). & Categorical \\ \hline
Gender & Gender of the student (Male, Female). & Categorical \\ \hline
Exam\_Score & Final exam score. & Numeric \\ \hline
\end{tabular}
\end{table}

\subsection*{B. SHAP Analysis}
\begin{figure}[htbp]
\centering
\includegraphics[width=0.9\columnwidth]{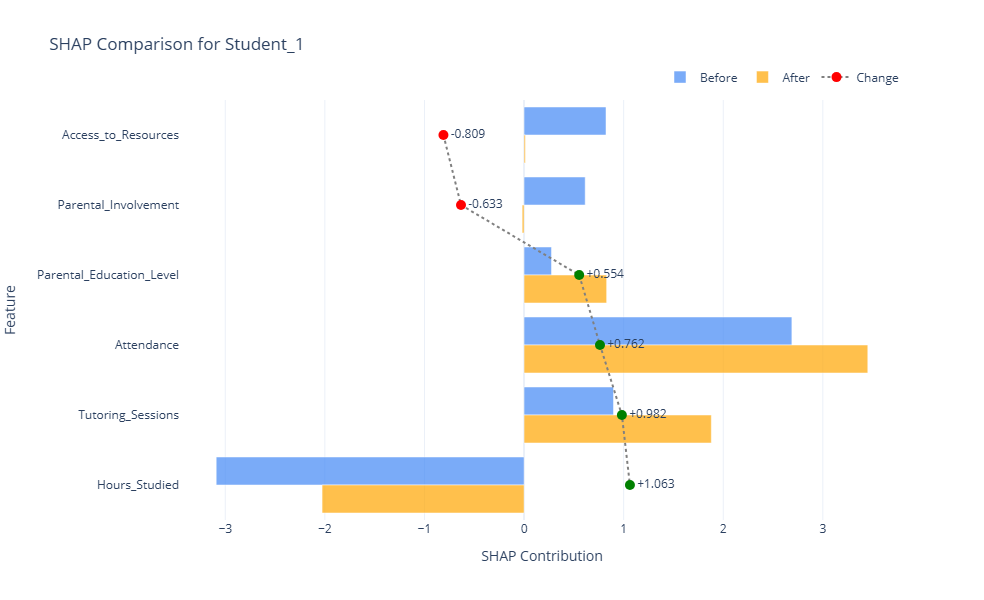}
\includegraphics[width=0.9\columnwidth]{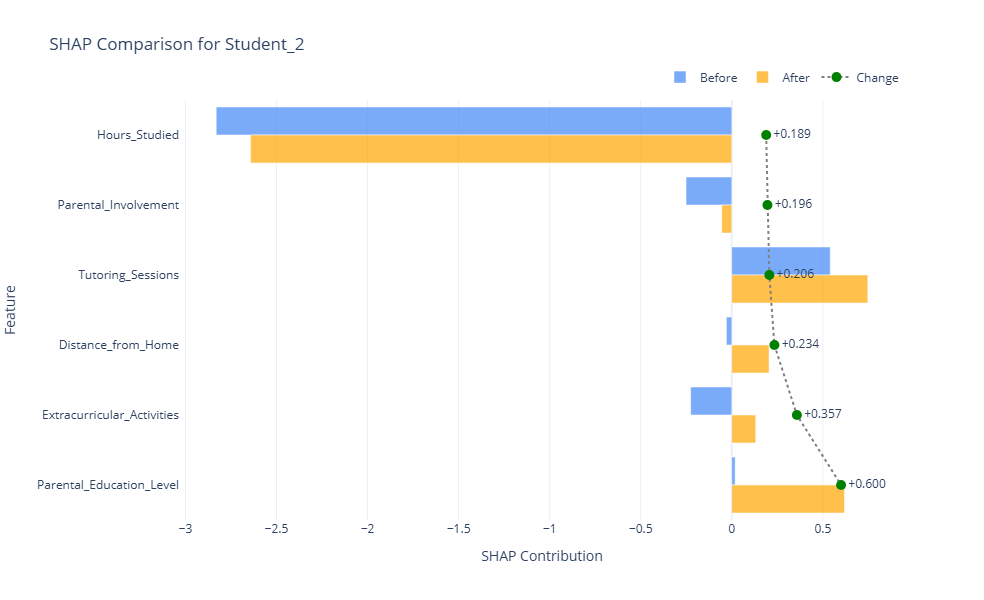}
\includegraphics[width=0.9\columnwidth]{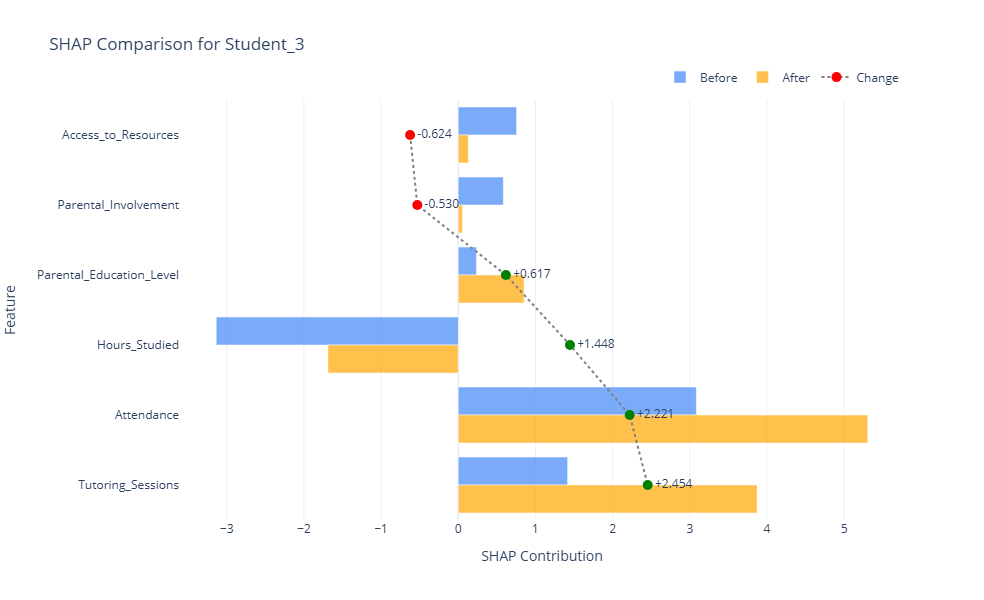}
\includegraphics[width=0.9\columnwidth]{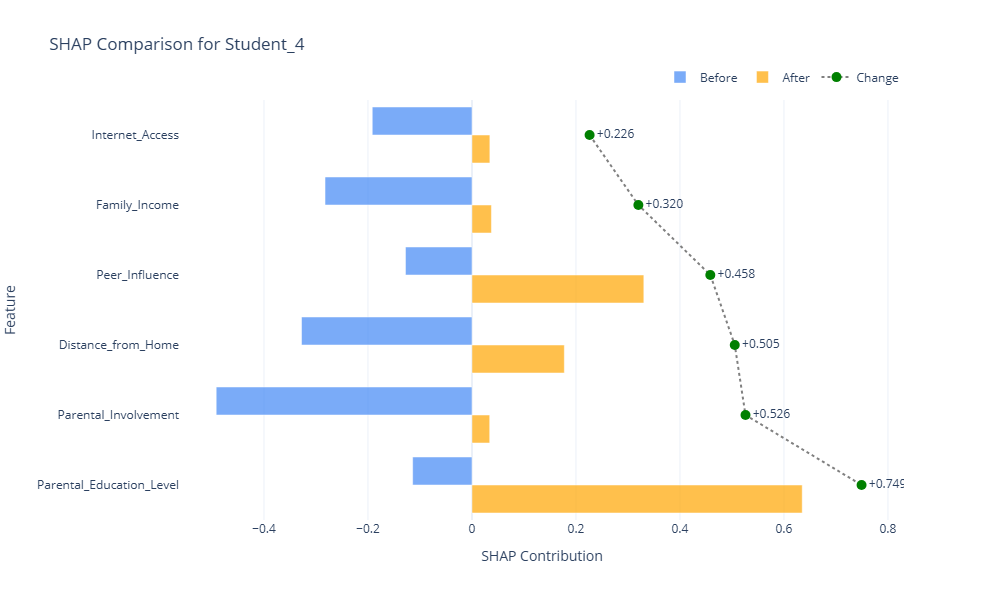}
\includegraphics[width=0.9\columnwidth]{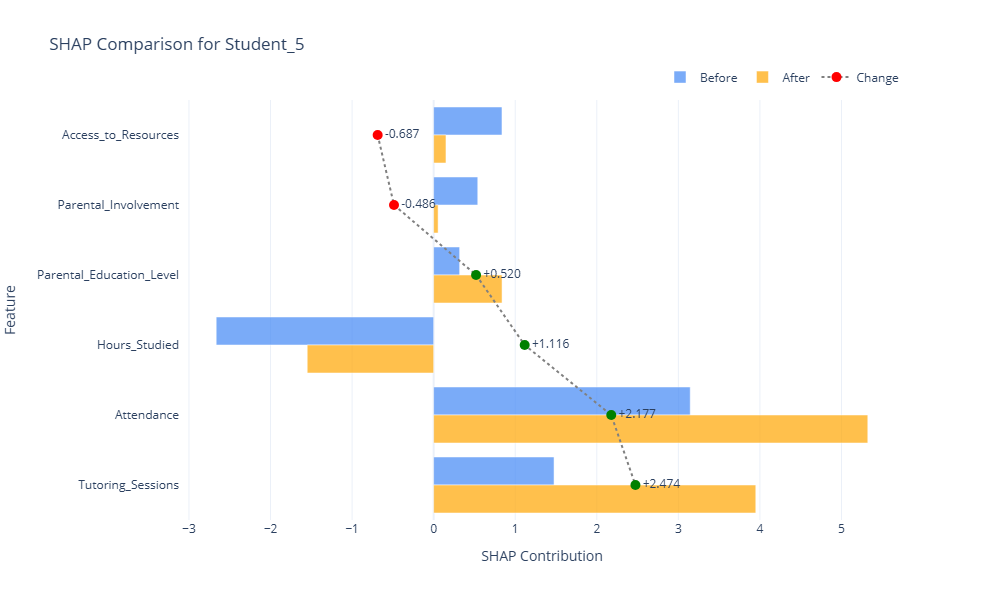}

\caption{SHAP feature importance after model retraining.}
\label{fig:shap_example}
\end{figure}

Figure~\ref{fig:shap_example} shows the SHAP summary plot after retraining, highlighting \texttt{Attendance}, \texttt{Hours\_Studied}, and \texttt{Tutoring\_Sessions} as top contributors to prediction changes.

\end{document}